\documentclass[journal]{IEEEtran}

\usepackage{amsmath,amssymb,amsfonts}
\usepackage{graphicx}
\usepackage[table]{xcolor} 
\usepackage{url}
\usepackage{cite}
\usepackage{tikz}
\usepackage{algcompatible}
\usepackage{algorithm}
\usepackage{multirow}
\usepackage{pifont}
\usepackage{xspace}

\usepackage{enumitem}
\usepackage[utf8]{inputenc}
\usepackage{xspace}
\newcommand{\eg}{e.g.\xspace}
\usepackage{microtype}
\usepackage{booktabs}
\usepackage[justification=centering]{caption}
\usepackage[table]{xcolor}
\usepackage{xcolor}
\usepackage{colortbl}
\usepackage{makecell}
\usepackage{tabularx}
\usepackage{array}
\usepackage{orcidlink}



\definecolor{highlightblue}{RGB}{235,235,255}

\title{Local-GS: Accelerating 3D Gaussian Splatting via Tile-Local Warp Coherence}

\author{
    Yang Luo\orcidlink{0009-0002-7124-2296}, 
    Yan Gong\orcidlink{0000-0002-3148-8286}, 
    Yongsheng Gao*\orcidlink{0000-0002-1555-8328}, 
    Jie Zhao\orcidlink{0000-0002-6086-9387},~\IEEEmembership{Senior Member,~IEEE}, 
    Xinyu Zhang\orcidlink{0000-0003-0034-9037},~\IEEEmembership{Member,~IEEE}, 
    and Huaping Liu\orcidlink{0000-0002-4042-6044},~\IEEEmembership{Fellow,~IEEE}%

\thanks{This work was supported by the National Science and Technology Major Project (Grant No. 2025ZD1603200) and the National Outstanding Youth Science Fund of the National Natural Science Foundation of China (Grant No. 52025054). (\textit{Corresponding author: Yongsheng Gao})}%
\thanks{Yang Luo, Yan Gong, Yongsheng Gao, and Jie Zhao are with the State Key Laboratory of Robotics and Systems, Harbin Institute of Technology, Harbin 150001, China (email: christoluo@outlook.com; gongyan2020@foxmail.com; gaoys@hit.edu.cn; jzhao@hit.edu.cn).}%
\thanks{Xinyu~Zhang is with the State Key Laboratory of Intelligent Green Vehicle and Mobility, the School of Vehicle and Mobility, Tsinghua University, Beijing 100084, China (email: xyzhang@tsinghua.edu.cn).}%
\thanks{Huaping~Liu is with the Department of Computer Science and Technology, Tsinghua University, Beijing 100084, China (email: hpliu@tsinghua.edu.cn).}%
}

\begin{document}

\maketitle

\begin{abstract}
3D Gaussian Splatting (3DGS) has significantly advanced real-time novel view synthesis by representing scenes as dense collections of anisotropic 3D Gaussian primitives. However, the irregular spatial distribution of Gaussians often leads to poor GPU utilization, as warp divergence and redundant computation degrade rendering performance. To address this, we present Local-GS, a warp-coherent rendering paradigm that, organizes Gaussian primitives with respect to SIMT (Single Instruction, Multiple Threads) execution boundaries rather than scene geometry. Specifically, we propose three warp-coherent stages: a hoisting stage that precomputes shared parameters at tile level, a culling stage that discards warps with no contribution, and a blending stage that replaces per-pixel branching with a uniform instruction stream. Across extensive benchmarks on multiple datasets, Local-GS improves efficiency without compromising quality. As a plug-and-play optimization, it provides additional performance gains to all tested baselines, culminating in a $7.76\times$ speedup on Deep Blending scenes. Code: \url{https://github.com/tilaba/Local-GS.git}.
\end{abstract}


\section{Introduction}
\label{sec:intro}
Real-time novel view synthesis is foundational to modern immersive applications such as AR/VR, digital twins, and autonomous driving simulation~\cite{mildenhall2020nerf, kung2025radarsplat, MemGS2025}. Among recent explicit 3D scene representations, 3D Gaussian Splatting (3DGS) represents scenes as anisotropic 3D Gaussian primitives, enabling high‑quality real‑time rendering~\cite{Kerbl20233DGS, RecentAdvanceWu2024, 3DGSSurvey2025}. It has since been extended to dynamic scene modeling~\cite{wu20244d, yang2024deformable} and large-scale city reconstruction~\cite{liu2025citygo}. However, its rendering efficiency degrades sharply as scene complexity and the number of primitives grows into the tens of millions~\cite{zhao2025scaling, Octree-GS2025}. In the standard rasterization pipeline, each pixel accumulates and blends projected Gaussians independently, regardless of their negligible contributions to pixel color~\cite{Kerbl20233DGS}. This indiscriminate processing induces warp divergence, unnecessary memory transactions, and redundant computations on modern GPUs. The problem becomes more pronounced in scenes with tens of millions of overlapping Gaussians (e.g., 10–40M)~\cite{RecentAdvanceWu2024, FlashGS2025, tian2025flexgaussian}.

To tackle these issues, existing methods predominantly adopt several strategies, including early culling, model compression, workload scheduling, and hardware-specific optimizations. Early culling algorithms~\cite{wang2024adr, speedy-splat} remove non-contributing Gaussians at the block level via tight bounding boxes or radius circles. We make a key observation: block‑level culling leaves intra‑block warp divergence untouched. Even after tile‑based culling, a non‑negligible fraction of warps inside a tile block still fetch and process Gaussians that ultimately contribute zero color to their assigned pixels. Primitive pruning strategies~\cite{fan2024lightgaussian, girish2024eagles, wang2026prune, lee2026optimized, huang2026seele, du2026mobile, compressionsurvey} reduce memory usage and computation, but often lead to a loss of high-frequency details and structural fidelity. In the rendering stage, throughput-oriented schedulers~\cite{FlashGS2025} and Tensor Core-based optimizations~\cite{liao2025tcgs} offer alternatives. However, both types of methods still exhibit limitations. Throughput-oriented schedulers suffer from warp divergence and extra blending computations, whereas Tensor-Core-based optimizations are tied to specific hardware and introduce numerical approximation errors. Most existing methods share a common focus: they select Gaussians based on scene geometry, rather than on how the hardware actually executes them. This motivates a central question: how can we achieve high‑performance tile‑based rasterization on general GPUs while avoiding warp divergence and redundant computation?

\begin{figure}
    \centering
    \includegraphics[trim={1.5cm 0.5cm 15cm 2cm}, clip, width=\linewidth]{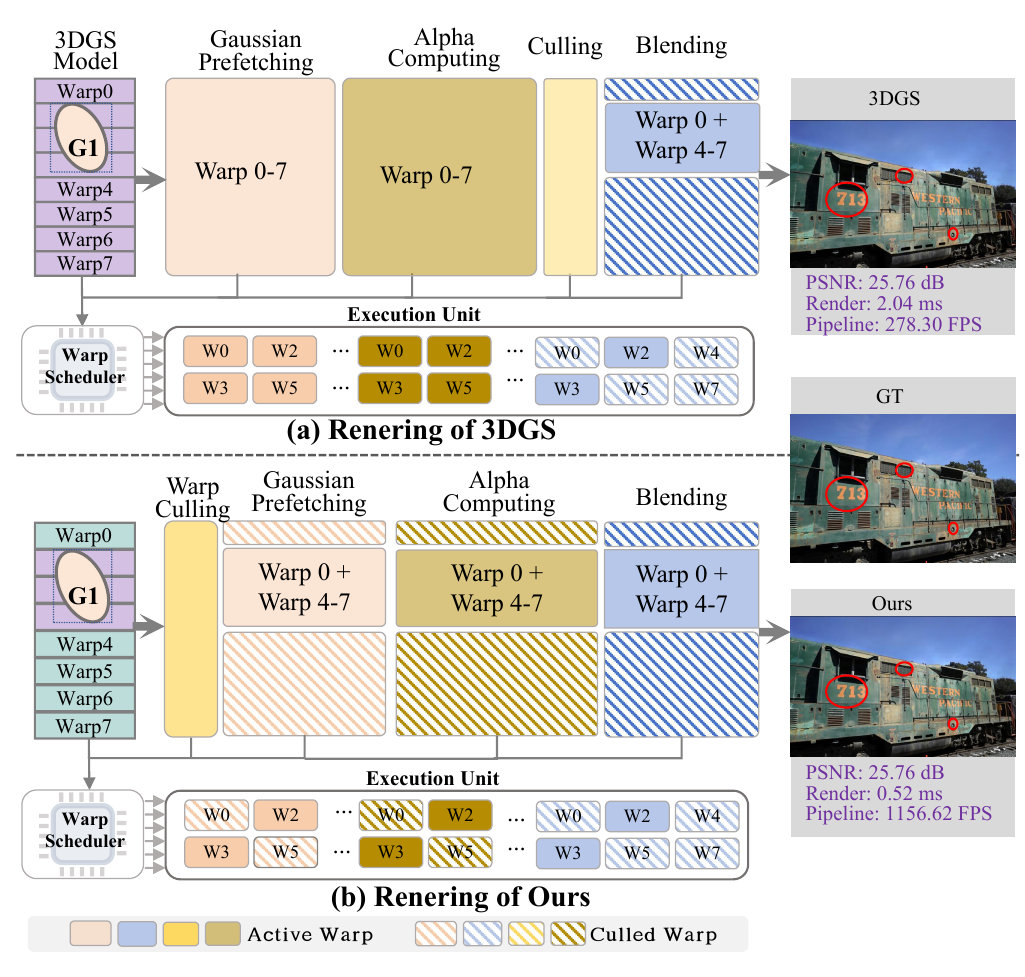}
    \captionsetup{
        font=small,
        labelfont=bf,
        justification=justified,
    }
    \caption{Rendering comparison. (a) \textbf{Standard 3DGS}  incurs heavy overhead because pixel culling is deferred until after alpha computation. (b) Following the conventional block-level culling, our Local-GS introduces warp-level culling before parameter prefetching. Maintaining a PSNR of 25.76 dB (averaged over the train scene in Tanks \& Templates~\cite{knapitsch2017tanks}), it reduces the rendering kernel time from 2.04 ms to 0.52 ms (also averaged over the train scene).}
    \label{fig:rendering_comparison}
\end{figure}

To answer this question, we shift the optimization focus from coarse-grained primitive culling to fine-grained GPU execution coherence. While block-level culling bounds the rendering workload to active block tiles, it fails to eliminate intra-block non‑contributing Gaussians before heavy computations. As illustrated in Fig.~\ref{fig:rendering_comparison}(a), all warps in the standard rendering kernel are forced to execute memory transactions and alpha computations before any per‑pixel culling takes effect. As culling is deferred to subsequent stages, even if a Gaussian ultimately yields no color contribution to the pixels covered by the warp, the primitive still incurs memory accesses and computations prior to culling. This indiscriminate execution model inflicts a non-trivial workload burden on the GPU, leading to warp divergence, unnecessary parameter loads, and instruction bubbles that degrade algorithm performance.

Building on these motivation, we propose \textbf{Local-GS}, a warp-coherent rendering paradigm that first restructures alpha computation and then reorganizes work around SIMT execution boundaries. We begin by hoisting all Gaussian-dependent terms into a tile-local 6D Gaussian representation, so that each pixel thread evaluates alpha via a unified dot-product form instead of repeatedly expanding the quadratic form. On top of this Gaussian-centric formulation, we further align execution with the GPU: as shown in Fig.~\ref{fig:rendering_comparison}(b), we shift conventional block-level culling to the intra-block warp level to skip inactive warps before Gaussian prefetching and alpha blending, and we rewrite the blending loop into a compact, branch-free arithmetic sequence. Together, these components provide substantial rendering speedups without compromising quality, through three contributions:

\begin{itemize}
\item We introduce \textbf{Local-GS}, a warp-coherent execution model that reduces SIMT divergence through tile-local hoisting.

\item We reformulate alpha computation into a \textbf{tile-local Gaussian representation} by hoisting
Gaussian-dependent terms into a shared 6D vector, so that per-pixel evaluation reduces to a unified
dot-product form with substantially reduced arithmetic cost.

\item We develop \textbf{warp-level culling} and \textbf{warp-coherent blending} scheme that align spatial visibility
and early termination with warp granularity, producing uniform SIMT execution orthogonal to
existing culling, pruning, and compression methods.
\end{itemize}

As a purely kernel-level optimization, Local‑GS can be seamlessly integrated into existing rendering pipelines without altering the underlying scene representation or relying on specialized hardware. On the Deep‑Blending dataset, when combined with FlashGS, our method achieves a cumulative \(7.76\times\) speedup over vanilla 3DGS on an RTX 4080 GPU, with zero loss in rendering quality. Moreover, Local‑GS is orthogonal to culling techniques such as Speedy‑Splat, 
and this orthogonality holds across different GPU architectures; when used alongside it, our method yields additional speedup while preserving visual fidelity.

\begin{figure}
    \centering
    \includegraphics[trim={3cm 7.5cm 9cm 2.5cm}, clip, width=\linewidth]{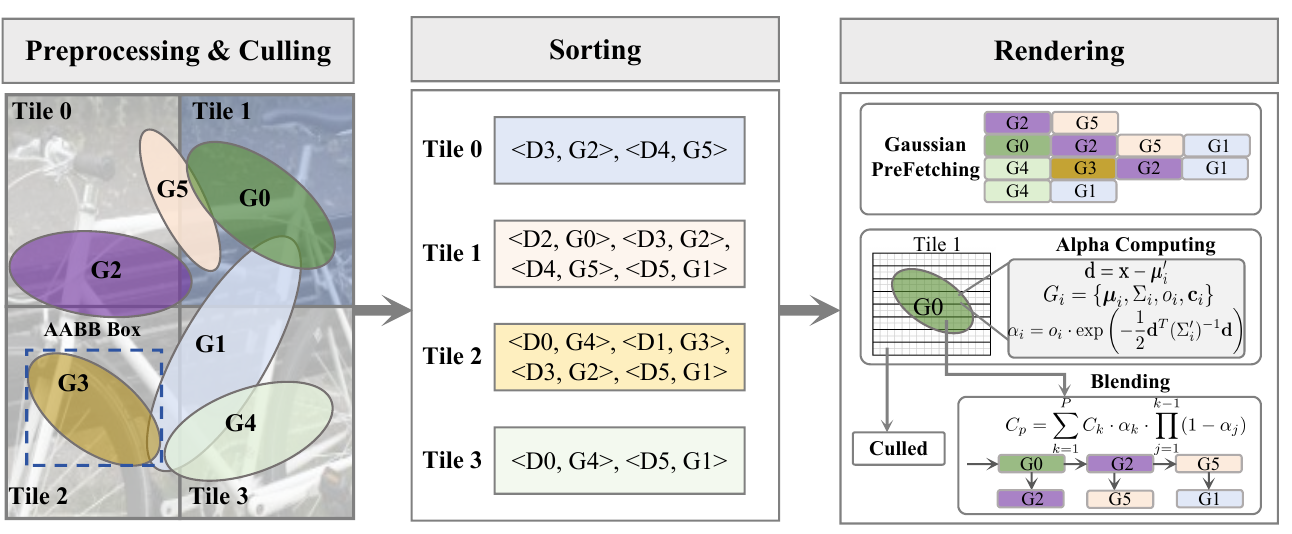}
    \captionsetup{
        font=small,
        labelfont=bf,
        justification=justified,
    }
    \caption{The rasterization pipeline of 3DGS. The execution consists of three sequential stages: preprocessing, sorting, and rendering.}
    \label{fig:pipeline_plot}
\end{figure}

\begin{figure*}[t]
    \centering
    \includegraphics[trim={1cm 0.0cm 1.5cm 0.0cm}, clip, width=1\linewidth]{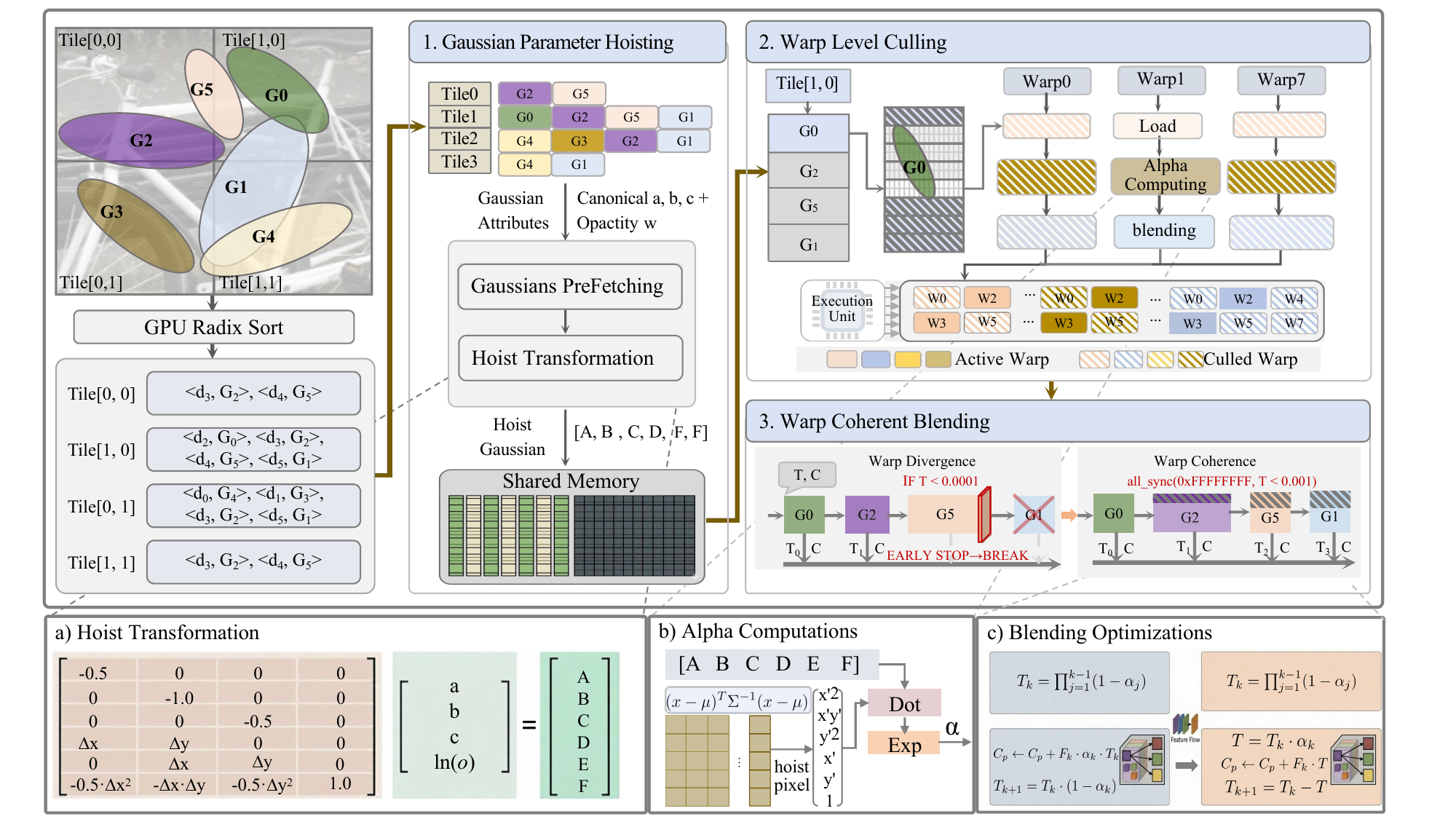}
    \captionsetup{
        font=small,
        labelfont=bf,
        justification=justified,
    }
    \caption{Overall architecture of Local-GS. The framework contains three key modules. (1) Gaussian Parameter Hoisting: It decouples tile-shared attributes and pre-computes them via a hoist transformation. Results are stored in shared memory for tile-wide reuse, eliminating redundant per-pixel workload. (2) Warp-Level Culling: Fine-grained culling at the warp granularity improves compute utilization. (3) Warp-Coherent Blending: Rasterization is restructured into a SIMT-aligned flow. Only pixel-dependent dot products are accumulated in the blending loop, reducing warp divergence.
Bottom panels detail (a) Hoist Transformation, (b) alpha computation, and (c) blending loop optimizations.}
    \label{fig:Overview of Framework}
\end{figure*}

\section{Related Work}
\label{sec:related}

\subsection{3D Gaussian Splatting}

As an explicit alternative to neural volumetric representations such as NeRF, 3D Gaussian Splatting (3DGS)~\cite{Kerbl20233DGS} models 3D scenes using a dense collection of anisotropic Gaussian primitives to achieve real-time novel view synthesis. Specifically, each Gaussian is characterized by its mean position $\mathbf{x}$, an opacity value $\alpha$, a covariance matrix $\mathbf{\Sigma}$ decomposed into scaling $\mathbf{S}$ and rotation $\mathbf{R}$, and view-dependent color parameterized by spherical harmonics ($\text{SH}$) coefficients. These primitives are rendered through a rasterization pipeline consisting of geometric preprocessing, depth sorting with duplication, and a tile-based rendering kernel responsible for alpha computation and blending.

As illustrated in Fig.~\ref{fig:pipeline_plot}, the pipeline first projects each 3D Gaussian onto the 2D screen space to compute its 2D mean and covariance. These attributes are then leveraged to determine a circular bounding radius or an axis-aligned bounding box (AABB) for block-level tile culling of Gaussian primitives. A parallel prefix scan then duplicates the surviving primitives across intersecting tiles into 64-bit key-value pairs, which are sorted by GPU Radix Sort to produce front-to-back index ranges (tile ranges) for each tile. The subsequent CUDA rendering kernel executes the Gaussian parameter prefetching, alpha computation and blending, which accounts for 55\%--85\% of the whole pipeline~\cite{Kerbl20233DGS, speedy-splat, FlashGS2025}. At the implementation level, the kernel maps a 16$\times$16 thread block to a single tile, mapping  each GPU thread to an individual pixel. To hide global memory latency, threads execute collaborative prefetching, loading serialized batches of Gaussian attributes into low-latency shared memory before independently traversing the list to compute alpha blending.

\subsection{Acceleration of 3D GS}

Recent acceleration methods for 3DGS improve the rendering throughput of the rasterization pipeline from four main perspectives: early culling~\cite{speedy-splat, wang2024adr, Peietal2025MICRO, zoomers2025nvgs}, model pruning and compression~\cite{ren2025fastgs, fang2024minisplatting, Compressed3DGS2024, Visibility2026, FeedforwardCompression,fan2024lightgaussian, HAC2015}, pipeline restructuring~\cite{FlashGS2025, Gui2024Balanced3G, 3DGStreamin2025}, and hardware-specific optimization~\cite{liao2025tcgs, li2026gemm, li2026accelerating}. Foremost among these, early culling strategies aim to filter out non-contributing Gaussians prior to the rendering stage by employing tighter bounds. For instance, AdR-Gaussian~\cite{wang2024adr} introduces an adaptive radius determined by minimum opacity thresholds to cull low-contribution ellipses. Meanwhile, Speedy-Splat~\cite{speedy-splat} focuses on tightening the 2D bounding boxes of projected Gaussians, performing a subsequent one-dimensional directional scan to determine exact tile intersections and eliminate redundant corner-tile processing. Despite the improved performance, rendering remains a major bottleneck in the overall pipeline. Another line of methods adopts model pruning and compression strategies, which reduce memory footprint and memory access overhead by modifying the underlying scene representation. For example, Mini-Splatting~\cite{fang2024minisplatting} employs Gaussian binarization and sampling to reorganize the spatial positions of Gaussian, while Compressed 3DGS\cite{Compressed3DGS2024} utilizes vector quantization and low-bit encoding to compress attribute data. Nonetheless, aggressive pruning sacrifices accuracy, requiring prolonged fine-tuning or secondary training to recover rendering quality. Pipeline restructuring and scheduling have also been explored to improve the performance of 3DGS. For instance, FlashGS~\cite{FlashGS2025} introduces stream-based scheduling and fused kernels for high hardware utilization. However, such designs are coupled with customized memory layouts, limiting their extensibility. Lastly, hardware-specific optimizations leverage specialized units such as Tensor Core-based optimizations to achieve performance improvements. For instance, TC-GS~\cite{liao2025tcgs} adopts this approach. However, the tight coupling with specific GPU hardware, particularly Tensor Cores, restricts general deployment across different GPU platforms while also inducing accuracy loss due to low-precision computation.

Unlike prior methods that require trade-offs in rendering quality, flexibility, and hardware universality, our Local-GS introduces a pure kernel-level optimization. By shifting the optimization focus from macroscopic scene culling down to microscopic GPU SIMT execution boundaries, our method mitigates intra-block computational overhead and warp divergence to maximize hardware utilization. Additionally, Local-GS is orthogonal to and compatible with existing compression, pruning, and macroscopic culling pipelines, serving as an effective plug-and-play acceleration module for a wide range of 3DGS frameworks.

\begin{table*}[t]
\centering
\small
\captionsetup{
    font=small,
    labelfont=bf,
    justification=justified,
}
\caption{Algorithm and instruction complexity inside the alpha computation loop.}
\label{tab:loop_complexity_comparison}
\setlength{\tabcolsep}{14.0pt}
\renewcommand{\arraystretch}{1.3}
\begin{tabular}{lll}
\toprule
\textbf{Metric / Operation} & \textbf{Baseline 3D-GS} & \textbf{Ours (Local-GS)} \\ 
\midrule
Power Calculation & $-0.5(a\Delta x^2 + c\Delta y^2) - b\Delta x\Delta y$ & $\mathbf{V}_{\text{hoisted}}^T \cdot \mathbf{B}_{\text{local}}$ \\
Global Coordinate Fetches & Required (Dynamic $x, y$) & None (Hoisted Outside) \\
Memory Access Size & $6 \times \text{float}$  & $6 \times \text{float}$ \\
Hardware Instruction Mapping & Scattered \texttt{SUB}, \texttt{MUL}, \texttt{ADD} sequence & Native FMA Chain \\
\midrule
\rowcolor{highlightblue}
Intra-loop Arithmetic Ops & 3 SUB, 8 MUL, 1 ADD & 5 FMA (Dot Product) \\
\bottomrule
\end{tabular}
\end{table*}

\section{Method}
\label{sec:method}

\subsection{Overview}
\label{subsec:overview}

Our Local-GS method is motivated by three underlying microarchitectural limitations within the rendering stage of the 3DGS pipeline. First, the thread-centric model forces repetitive gaussian parameter calculations across neighboring pixels, leading to  redundant computation for shared primitive attributes. Second, coarse tile-level culling overlooks localized spatial distributions, causing intra-block workload imbalance.Third, the sequential accumulation of irregular primitives breaks execution uniformity and degrades SIMT throughput during blending.

To tackle the above bottlenecks, Local-GS enforces warp‑coherent, tile‑local execution through three complementary techniques (Fig.~\ref{fig:Overview of Framework}).
Tile-Hoisted Parameterization (Sec.~\ref{subsec:hoisting}) precomputes all pixel‑independent coefficients once per Gaussian in shared memory, reducing per‑pixel alpha computation to a dot product.
Warp‑Level Culling (Sec.~\ref{subsec:culling}) subdivides each tile into warp‑aligned strips and skips entire warps whose pixels are not covered by the current Gaussian, preventing wasted memory fetches and arithmetic.
Warp‑Coherent Blending (Sec.~\ref{subsec:blending}) replaces per‑pixel branching with warp‑uniform exit conditions and restructures the blending loop into a dense FMA stream.
Together, these mechanisms transform irregular, geometry‑dependent rasterization into a uniform, warp‑coherent execution flow.

\subsection{Gaussian Parameter Hoisting}
\label{subsec:hoisting}

In the standard 3DGS rasterization pipeline, alpha computation is performed independently by each pixel thread.
Every thread evaluates the identical Gaussian parameters---conic matrix coefficients and opacity---and combines them with its own pixel-specific displacement.
This pixel-centric model forces redundant computation of shared attributes across all threads that a Gaussian covers, wasting arithmetic throughput and memory bandwidth inside the hot blending loop.

To eliminate this redundancy, we draw inspiration from loop-invariant code motion in classical compiler optimization~\cite{aho2006compilers, allen1971catalogue}.
The key insight is that Gaussian parameters are \emph{pixel-independent}: once a Gaussian is projected onto a tile, its conic matrix and opacity are fixed for all pixels within that tile.
We therefore hoist all pixel-independent computation out of the per-pixel blending loop and precompute these parameters once per Gaussian during a cooperative prefetching phase.
This transforms the hot loop from a repeated quadratic expansion into a lightweight vector dot product. It simultaneously reduces arithmetic cost and makes all threads in a warp execute the same fixed‑length FMA chain for each Gaussian.

\subsubsection{Tile-Local Coordinate Reformulation}

We begin by shifting the reference frame from global screen space to the local coordinate system of each execution tile.
Let $\mathbf{T} = (T_x, T_y)^T$ be the global pixel coordinates of the tile's top-left corner.
For any pixel with global coordinates $\mathbf{x} = (x, y)^T$, its tile-local coordinates $\mathbf{x}' = (x', y')^T$ are:
\begin{equation}
\label{eq:tile-local}
\mathbf{x}' = (x - T_x,\; y - T_y).
\end{equation}
Similarly, the projected 2D center $\boldsymbol{\mu} = (\mu_x, \mu_y)^T$ of a Gaussian is expressed in tile-relative offsets:
\begin{equation}
\Delta_x = \mu_x - T_x, \quad \Delta_y = \mu_y - T_y.
\end{equation}
The displacement vector from the Gaussian center to the pixel then becomes purely local:
\begin{equation}
\tilde{\mathbf{x}} = \begin{bmatrix} \tilde{x} \\ \tilde{y} \end{bmatrix} = \begin{bmatrix} x' - \Delta_x \\ y' - \Delta_y \end{bmatrix}.
\end{equation} Intuitively, this change of variables simply says: instead of describing everything in global screen coordinates, each tile uses its own local (0,0) origin at the top‑left corner. All pixel positions and Gaussian centers are then expressed as small offsets within this tile, which keeps numerical values bounded and simplifies subsequent algebra.

Recall the standard alpha formulation in 3DGS~\cite{Kerbl20233DGS}:
\begin{equation}
\label{3ds_formulation}
\alpha = o \cdot \exp\!\left( -\frac{1}{2}\, \tilde{\mathbf{x}}^T \mathbf{\Sigma}^{-1} \tilde{\mathbf{x}} \right),
\end{equation}
where $o$ is the intrinsic opacity and the 2D inverse covariance (conic) matrix is:
\begin{equation}
\label{eq:conic_matrix}
\mathbf{\Sigma}^{-1} = \begin{bmatrix} a & b \\ b & c \end{bmatrix}.
\end{equation} Substituting the local displacement into Eq.~\eqref{3ds_formulation} and expanding yields a quadratic form entirely in tile-local coordinates.

\subsubsection{Canonical Polynomial Expansion and Hoist Transformation}

To absorb the opacity multiplication and isolate pixel-independent terms, we work in the log domain:
\begin{equation}
\ln(\alpha) = \ln(o) - \frac{1}{2}\Big[ a(x'-\Delta_x)^2 + 2b(x'-\Delta_x)(y'-\Delta_y) + c(y'-\Delta_y)^2 \Big].
\end{equation} Expanding and grouping terms by the local pixel basis $(x'^2, x'y', y'^2, x', y', 1)$ yields a canonical polynomial:
\begin{equation}
\label{eq:canonical}
\ln(\alpha) = A\,x'^2 + B\,x'y' + C\,y'^2 + D\,x' + E\,y' + F.
\end{equation} Conceptually, this is just a polynomial regrouping. For each Gaussian, we precompute how much it contributes to the six basis terms $(x'^2, x'y', y'^2, x', y', 1)$. Once these six coefficients (A–F) are known, any pixel in the tile can obtain its log‑opacity by a single weighted sum of these six basis values.

Matching coefficients gives a single linear operator that maps the Gaussian's intrinsic parameters to a 6D hoisted vector $\mathbf{V}_{\text{hoisted}} = [A, B, C, D, E, F]^T$:
\begin{equation}
\label{eq:hoist_transform}
\begin{bmatrix} A \\ B \\ C \\ D \\ E \\ F \end{bmatrix} = 
\begin{bmatrix} 
-0.5 & 0 & 0 & 0 \\ 
0 & -1.0 & 0 & 0 \\ 
0 & 0 & -0.5 & 0 \\ 
\Delta_x & \Delta_y & 0 & 0 \\ 
0 & \Delta_x & \Delta_y & 0 \\ 
-0.5\Delta_x^2 & -\Delta_x\Delta_y & -0.5\Delta_y^2 & 1.0 
\end{bmatrix}
\begin{bmatrix} a \\ b \\ c \\ \ln(o) \end{bmatrix}.
\end{equation}

This \textbf{Hoist Transformation} (Fig.~\ref{fig:Overview of Framework}(a)) is executed exactly once per Gaussian during the prefetching stage, before any pixel enters the blending loop.
The resulting vector $\mathbf{V}_{\text{hoisted}}$ compactly encodes all geometric and opacity information of the Gaussian in tile-local form.
At the hardware level, it is stored in shared memory as one \texttt{float4} (for $A, B, C, F$) and one \texttt{float2} (for $D, E$), enabling coalesced, aligned memory transactions without allocating a separate register for opacity.

\subsubsection{Per-Pixel Alpha as a Dot Product}

Once hoisted, per-pixel alpha evaluation decouples into two independent components: the precomputed Gaussian vector $\mathbf{V}_{\text{hoisted}}$ and a local pixel basis $\mathbf{B}_{\text{local}}$ that is fixed per thread and can be kept in registers:
\begin{equation}
\mathbf{B}_{\text{local}} = [x'^2,\; x'y',\; y'^2,\; x',\; y',\; 1]^T.
\end{equation}

Inside the blending hot loop, alpha computation reduces to a single dot product followed by an exponential:
\begin{equation}
\alpha = \exp\!\left( \mathbf{V}_{\text{hoisted}}^T \cdot \mathbf{B}_{\text{local}} \right).
\end{equation}

This formulation provides two critical benefits for warp-coherent execution.
First, it reduces redundant arithmetic: the quadratic expansion, which in the baseline requires a scattered sequence of subtractions, multiplications, and additions per Gaussian-pixel pair, is replaced by $5$ multiplications and $5$ additions that map directly onto $5$ hardware FMA instructions (Table~\ref{tab:loop_complexity_comparison}).
Second, and more fundamentally, every thread in a warp executes the \emph{identical} dot product instruction sequence with no conditional masking.
This achieves instruction-level uniformity across the warp, allowing the warp scheduler to issue at peak SIMT throughput throughout the entire blending loop.

\subsubsection{Precision Compensation}

Beyond arithmetic reduction, the tile-local reparameterization intrinsically suppresses catastrophic cancellation---a well-known hazard when subtracting large, nearly equal floating-point arithmetic~\cite{ieee7542008}.

In global coordinates, high-resolution rendering (e.g., 4K) pushes $x$ and $\mu_x$ to values around $4000$.
Evaluating the quadratic form $f(x)=x^2-2\mu_x x+\mu_x^2$ directly in this range yields an absolute error bound:
\begin{equation}
\begin{aligned}
E_{\text{global}} &\le \big| f(x) - \text{fl}(x^2 - 2\mu_x x + \mu_x^2) \big| \\
&\le \big( x^2 + 2|\mu_x x| + \mu_x^2 \big) \cdot \epsilon_{\text{mach}} \\
&\approx 4 \times 4000^2 \times 5.96 \times 10^{-8} \approx 3.81,
\end{aligned}
\end{equation}
where $\epsilon_{\text{mach}} = 2^{-24}$ for \texttt{float32}. When the true sub-pixel displacement is minuscule (e.g., $x-\mu_x = 0.2$, giving a theoretical quadratic value of only $0.04$), this error bound completely overwhelms the signal, producing visible banding and aliasing in high-frequency regions.

In our tile-local frame, all coordinates are bounded by the tile dimension.
For a standard $16 \times 16$ tile, $x'$ and $\Delta_x$ lie within $[-16, 32]$, reducing the error bound by four orders of magnitude:
\begin{equation}
\begin{aligned}
E_{\text{local}} &\le \big( x'^2 + 2|\Delta_x x'| + \Delta_x^2 \big) \cdot \epsilon_{\text{mach}} \\
&\le 4 \times 32^2 \times 5.96 \times 10^{-8} \approx 2.44 \times 10^{-4}.
\end{aligned}
\end{equation}
In simpler terms, evaluating the quadratic in global coordinates may involve subtracting numbers around 4000 that differ only in the first decimal place, which is numerically fragile in fp32. By switching to tile‑local coordinates, all values stay within roughly [-16,\,32], so we avoid subtracting large nearly‑equal numbers. This alone is enough to eliminate visible banding and recover the baseline 3DGS quality.

\subsubsection{Cooperative Staging and Shared Memory Layout}

Immediately before alpha computation, all threads in a tile block cooperatively load the hoisted 6D parameter vectors and associated view-dependent colors from global memory into shared memory via coalesced transactions (Gaussian Prefetching in Fig.~\ref{fig:Overview of Framework}).
During blending, threads read these parameters directly from shared memory, minimizing global memory traffic.
This cooperative layout is designed to match the warp-aligned execution pattern: each warp accesses its own subset of Gaussians determined by warp-level granularity (Sec.~\ref{subsec:culling}), ensuring that shared memory bandwidth is consumed only by active warps.

\subsection{Warp-Level Culling}
\label{subsec:culling}

\subsubsection{Warp-Level Tile Partitioning}
To explicitly map spatial sparsity to the GPU hardware execution hierarchy, our framework abandons the traditional pixel-centric independent traversal paradigm. As illustrated in the overall architecture (Fig.~\ref{fig:Overview of Framework}), each $16 \times 16$ pixel tile is internally subdivided into $W_{\text{warp}} = 8$ independent $16 \times 2$ pixel sub-regions. This fine-grained geometric partitioning aligns the spatial visibility culling granularity with the hardware's SIMT execution units, where each $16 \times 2$ sub-tile maps to a single 32-thread hardware warp.

\begin{figure}
    \centering
    \includegraphics[trim={3.5cm 3cm 16cm 9cm}, clip, width=\linewidth]{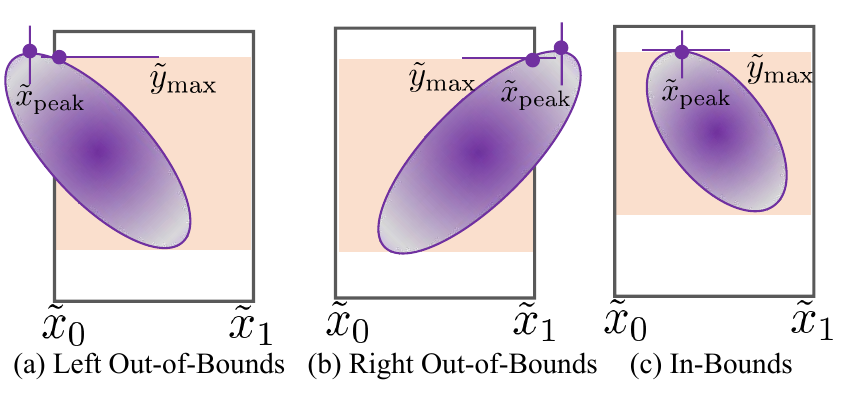}
    \captionsetup{
        font=small,
        labelfont=bf,
        justification=justified,
    }
    \caption{Three bounding box culling cases on a 16‑pixel wide, 2‑pixel high warp strip. Cases are decided by $\tilde{x}_{\text{peak}}$ relative to the strip boundaries $[\tilde{x}_0, \tilde{x}_1]$.}
    \label{fig:case_A_B_C}
\end{figure}

\subsubsection{Tight Bounding Box}
Traditional 3DGS frameworks utilize a standard square Axis-Aligned Bounding Box (AABB) derived from block-level coarse culling, which introduces significant redundant footprint for anisotropic Gaussians. To address this, we introduce a compact bounding box (\textbf{Tight Box}) constrained within a 1D column strip for warp granularity.

Following the standard 2D Gaussian formulation and the conic matrix elements $a, b, c$ defined in Eq.~\eqref{eq:conic_matrix}, the spatial footprint boundary under the global opacity threshold $\epsilon = 1/255$ forms a conic inequality $\tilde{\mathbf{x}}^T \mathbf{\Sigma}^{-1} \tilde{\mathbf{x}} \le \tau$, where $\tau = 2\ln(255 \cdot o)$ represents the dynamic threshold adjusted by the Gaussian's inherent opacity $o$. Expanding this yields the continuous boundary ellipse equation:
\begin{equation}
a\tilde{x}^2 + 2b\tilde{x}\tilde{y} + c\tilde{y}^2 = \tau.
\end{equation}
When a Gaussian primitive is projected onto the current tile column, its horizontal activity span is bounded by the left and right global screen coordinates $[X_0, X_1]$ of that column. The center-relative horizontal displacements are thus tightly clamped to the interval $\tilde{x} \in [\tilde{x}_0, \tilde{x}_1]$, where:
\begin{equation}
\tilde{x}_0 = X_0 - \mu_x, \quad \tilde{x}_1 = X_1 - \mu_x.
\end{equation}
To determine a tight bounding box within this 16-pixel wide strip, we rewrite the boundary ellipse into a quadratic form with respect to the vertical center-relative displacement $\tilde{y}$:
\begin{equation}
c\tilde{y}^2 + 2b\tilde{x}\tilde{y} + (a\tilde{x}^2 - \tau) = 0.
\end{equation}
By analyzing this quadratic profile, the unconstrained maximum vertical radius $d_y$ of the ellipse is derived as:
\begin{equation}
d_y = \sqrt{\frac{a\tau}{ac-b^2}}.
\end{equation}
Leveraging the positive-definite property of the conic matrix ($a > 0$), we evaluate whether the vertical peak of the ellipse falls inside or outside the current horizontal clipping interval $[\tilde{x}_0, \tilde{x}_1]$. This is implemented via a division-free condition check optimized for hardware execution:

\small
\vspace{0.5em}
\noindent\textbf{Case A (Left Out-of-Bounds):} If $-b \cdot d_y < a \cdot \tilde{x}_0$, the vertical peak vertex falls to the left of the column, as illustrated in Fig.~\ref{fig:case_A_B_C}(a). The profile behaves monotonically within the strip, meaning the maximum vertical displacement $\tilde{y}_{\max}$ must occur at the left boundary $\tilde{x}_0$:
\begin{equation}
\tilde{y}_{\max} = \frac{-b\tilde{x}_0 + \sqrt{c\tau - (ac-b^2)\tilde{x}_0^2}}{c}.
\end{equation}

\noindent\textbf{Case B (Right Out-of-Bounds):} If $-b \cdot d_y > a \cdot \tilde{x}_1$, the peak vertex falls to the right of the column (see Fig.~\ref{fig:case_A_B_C}(b)), and the maximum displacement within the strip is achieved at the right boundary $\tilde{x}_1$:
\begin{equation}
\tilde{y}_{\max} = \frac{-b\tilde{x}_1 + \sqrt{c\tau - (ac-b^2)\tilde{x}_1^2}}{c}.
\end{equation}

\noindent\textbf{Case C (In-Bounds):} If $a \cdot \tilde{x}_0 \le -b \cdot d_y \le a \cdot \tilde{x}_1$, the peak vertex is enclosed within the tile column (Fig.~\ref{fig:case_A_B_C}(c)). The maximum vertical displacement inside the column natively equals the unconstrained radius, i.e., $\tilde{y}_{\max} = d_y$.
\vspace{0.5em}
\normalsize

Symmetrically, by employing the identical conditional logic with reversed signs for the linear terms, the minimum vertical boundary displacement $\tilde{y}_{\min}$ within the strip is resolved. This evaluation yields a compact vertical span $[y_{\min}, y_{\max}]$ within the tile column, mapped to global screen coordinates via:
\begin{equation}
y_{\min} = \mu_y + \tilde{y}_{\min}, \quad y_{\max} = \mu_y + \tilde{y}_{\max}.
\end{equation}

\subsubsection{Warp Visibility Mask Construction}
Once the continuous vertical span is obtained, it is intersected with the vertical boundary of the current tile to yield the valid range $[y_{\text{start}}, y_{\text{end}}] = [\max(y_{\text{tile}}^{\text{min}}, \lceil y_{\min} \rceil), \min(y_{\text{tile}}^{\text{max}}, \lfloor y_{\max} \rfloor)]$. If $y_{\text{start}} \le y_{\text{end}}$, we map the span to discrete warp indices using fast hardware bitwise shift operations:
\begin{equation}
\begin{aligned}
w_{\text{start}} &= (y_{\text{start}} - y_{\text{tile}}^{\text{min}}) \gg 1, \\
w_{\text{end}}   &= (y_{\text{end}} - y_{\text{tile}}^{\text{min}}) \gg 1,
\end{aligned}
\end{equation}
where $y_{\text{tile}}^{\text{min}}$ represents the global vertical pixel coordinate of the top boundary of the current tile. Since each warp covers exactly 2 pixel rows, a single right-shift bitwise operation ($\gg 1$) achieves efficient discretization.

Accordingly, the visibility of each Gaussian primitive relative to the current tile column is implicitly encoded into an 8-bit unsigned integer mask $\mathbf{M}_i \in \{0, 1\}^8$:
\begin{equation}
\mathbf{M}_i = \left( (1 \ll (w_{\text{end}} - w_{\text{start}} + 1)) - 1 \right) \ll w_{\text{start}}.
\end{equation}
If $y_{\text{start}} > y_{\text{end}}$, the primitive is completely outside the tile, and $\mathbf{M}_i = 0$. Each bit of $\mathbf{M}_i$ explicitly denotes whether the primitive yields a spatial contribution to the corresponding $16 \times 2$ warp sub-region, transforming continuous geometric clipping into a hardware-friendly discrete warp mask.

\subsubsection{Instruction Execution Flow}
As illustrated in the execution path of Module 2 (Fig.~\ref{fig:Overview of Framework}), our warp-level culling optimizes the micro-architectural interaction between the hardware warp scheduler and Execution Units.  It replaces traditional predicated idling with uniform branches. Standard pixel-centric 3DGS rendering relies on predicated execution when a Gaussian primitive partially intersects a tile. This forces the warp scheduler to issue the complete \texttt{Load}-\texttt{Alpha}-\texttt{Blending} instruction chain for all 32 threads within a warp, leaving execution units idling and introducing significant instruction bubbles that degrade instructions per cycle (IPC). Building on this analysis, we leverage a pre-computed 8-bit visibility mask $\mathbf{M}_i$ as a front-end jump barrier. For culled warps ($\mathbf{M}_i[w] = 0$), the absolute convergence of all 32 threads instantly triggers a hardware uniform branch, allowing the warp scheduler to bypass the downstream instruction sequence with zero issue overhead. Hardware execution units are thus utilized exclusively for active warps ($\mathbf{M}_i[w] = 1$), effectively mitigating pipeline bubbles and maximizing GPU compute density.

\begin{table*}[t]
\captionsetup{
    font=small,
    labelfont=bf,
    justification=justified,
}
\setlength{\tabcolsep}{0.50pt}
\renewcommand{\arraystretch}{1.3}
\centering
\caption{End-to-end rendering performance and quality comparison on RTX 4080. Values in parentheses in the FPS column denote the speedup factor relative to vanilla 3DGS~\cite{Kerbl20233DGS}.}
\label{tab:pipeline_perf}
\begin{tabular}{l|cccc|cccc|cccc}
\noalign{\hrule height 1.2pt}
 & \multicolumn{4}{c|}{Tanks \& Templates} & \multicolumn{4}{c|}{Deep Blending} & \multicolumn{4}{c}{Mip-NeRF360} \\
Method & FPS $\uparrow$ & PSNR $\uparrow$ & SSIM $\uparrow$ & LPIPS $\downarrow$ & FPS $\uparrow$ & PSNR $\uparrow$ & SSIM $\uparrow$ & LPIPS $\downarrow$ & FPS $\uparrow$ & PSNR $\uparrow$ & SSIM $\uparrow$ & LPIPS $\downarrow$ \\ \hline
3DGS\cite{Kerbl20233DGS} (2023)                  & 287.12  & 26.657 & 0.892 & 0.151 & 206.37  & 36.549 & 0.951 & 0.195 & 192.79 & 29.420 & 0.851 & 0.209 \\
\rowcolor{highlightblue} 3DGS (+Local-GS)        & 466.84 (1.63$\times$)  & 26.657 & 0.892 & 0.151 & 356.37 (1.73$\times$)  & 36.549 & 0.951 & 0.195 & 315.42 (1.64$\times$) & 29.420 & 0.851 & 0.209 \\
FastGS\cite{ren2025fastgs} (2025)                & 640.61  & 26.248 & 0.890 & 0.151 & 732.80  & 36.133 & 0.949 & 0.193 & 590.55 & 29.360 & 0.851 & 0.208 \\
\rowcolor{highlightblue} FastGS (+Local-GS)      & 1114.80 (3.88$\times$) & 26.248 & 0.890 & 0.151 & 1168.88 (5.66$\times$) & 36.132 & 0.949 & 0.193 & 956.76 (4.96$\times$) & 29.360 & 0.851 & 0.208 \\
AdR-Gaussian\cite{wang2024adr} (2024)            & 639.00  & 26.583 & 0.892 & 0.151 & 644.57  & 36.484 & 0.950 & 0.195 & 504.67 & 29.382 & 0.851 & 0.208 \\
\rowcolor{highlightblue} AdR-Gaussian (+Local-GS)& 1001.83 (3.49$\times$) & 26.582 & 0.892 & 0.151 & 974.22 (4.72$\times$) & 36.484 & 0.950 & 0.194 & 745.79 (3.87$\times$) & 29.382 & 0.851 & 0.208 \\
Speedy-Splat\cite{speedy-splat} (2025)           & 630.89  & 26.656 & 0.892 & 0.151 & 652.71  & 36.549 & 0.951 & 0.195 & 529.47 & 29.420 & 0.851 & 0.209 \\
Speedy-Splat (+TC-GS)\cite{liao2025tcgs}         & 715.48 (2.49$\times$)  & 26.635 & 0.892 & 0.150 & 711.04 (3.45$\times$)  & 36.515 & 0.950 & 0.194 & 571.50 (2.96$\times$) & 29.407 & 0.851 & 0.208 \\
\rowcolor{highlightblue} Speedy-Splat (+Local-GS)& 1058.07 (3.69$\times$)  & 26.656 & 0.892 & 0.151 & 1035.23 (5.02$\times$)  & 36.549 & 0.951 & 0.195 & 829.03 (4.30$\times$) & 29.420 & 0.851 & 0.209 \\
FlashGS\cite{FlashGS2025} (2025)                     & 756.79  & 26.657 & 0.892 & 0.151 & 1245.14 & 36.549 & 0.951 & 0.195 & 344.40 & 29.420 & 0.851 & 0.209 \\
\rowcolor{highlightblue} FlashGS (+Local-GS)    & 822.41 (2.86$\times$)  & 26.657 & 0.892 & 0.151 & 1601.77 (7.76$\times$) & 36.549 & 0.951 & 0.195 & 360.07 (1.87$\times$) & 29.420 & 0.851 & 0.209 \\
\noalign{\hrule height 1.2pt}
\end{tabular}
\end{table*}

\subsection{Warp-Coherent Blending}
\label{subsec:blending}

We eliminate the two remaining sources of SIMT divergence in the blending loop (per-pixel early termination and opacity-threshold branching) and restructure the arithmetic into a dense FMA stream.

\subsubsection{Divergence Elimination}
\label{subsec:df_alpha}

The vanilla baseline evaluates the termination condition $T \cdot (1 - \alpha) < 0.0001$ on a per-pixel basis using a conditional branch, which severely increases warp divergence. To eliminate this bottleneck, we lift early termination to warp-aligned granularity (see Algorithm~\ref{alg:conditional-blending}). Specifically, the warp uniformly evaluates the collective status via the \texttt{\_\_all\_sync} intrinsic, defined as
\begin{equation}
\Phi_{\text{stop}}(\mathbf{T}) = \bigwedge_{i=1}^{32} (T_i < 0.0001),
\end{equation}

and skips subsequent blending only when all threads within the warp reach consensus. Due to the tight $16 \times 2$ warp-level culling, pixels within the same warp naturally exhibit high spatial coherence, aligning their transmittance attenuation and avoiding single-thread bottlenecks. Furthermore, the opacity computation is reformulated to be entirely branch-free: $\alpha$ is derived by exponentiating the hoisted log-power and clamping the result to $[0, 0.99]$, which inherently decays to zero below the visibility threshold in the log domain.

\begin{algorithm}[t]
\centering
\footnotesize
\setlength{\tabcolsep}{2pt}
\caption{Comparison of blending loop execution}
\label{alg:conditional-blending}
\vspace{0.3em}
\begin{tabular}{@{} p{0.46\linewidth} @{\hspace{0.02\linewidth}} p{0.46\linewidth} @{}}
\multicolumn{2}{@{}l@{}}{\textbf{Baseline Blending} \hspace{0.24\linewidth} \textbf{Our Optimized Blending}} \\
\midrule
\begin{minipage}[t]{\linewidth}
\ttfamily
\setlength{\parindent}{0pt}
\textbf{foreach} Gaussian $G'$ \textbf{do}\\
\quad $test\_T \leftarrow T \times (1 - \alpha)$;\\
\quad \textbf{if} $test\_T < 0.0001$ \textbf{then}\\
\quad\quad $done \leftarrow \text{true}$;\\
\quad\quad \textbf{break};\\
\quad \textbf{end}\\
\quad \textbf{foreach} channel $ch \in [0, N)$ \textbf{do}\\
\quad\quad $C[ch] \leftarrow C[ch] + f[ch] \times \alpha \times T$;\\
\quad \textbf{end}\\
\quad $T \leftarrow test\_T$;\\
\textbf{end}
\end{minipage}
&
\begin{minipage}[t]{\linewidth}
\ttfamily
\setlength{\parindent}{0pt}
\textbf{if} $\mathtt{\_\_all\_sync}(\mathtt{0xFFFFFFFF},
\ T \\< 0.0001)$ \textbf{then}\\
\quad \textbf{continue};\\
\textbf{end}\\
\textbf{foreach} Gaussian $G'$ \textbf{do}\\
\quad $w \leftarrow \alpha \times T$;\\
\quad \textbf{foreach} channel $ch \in [0, N)$ \textbf{do}\\
\quad\quad $C[ch] \leftarrow C[ch] + f[ch] \times w$;\\
\quad \textbf{end}\\
\quad $T \leftarrow T - w$;\\
\textbf{end}
\end{minipage}
\\
\end{tabular}
\end{algorithm}

\subsubsection{Blending Optimization}
\label{subsec:blending_opt}

To further maximize throughput, we optimize the inner blending arithmetic. As shown in Algorithm~\ref{alg:conditional-blending}, the pseudocode compares the execution path per Gaussian for the baseline and our design (assuming channels $N=3$ for RGB).

In the baseline, two conditional checks remain inside the loop: one against a fixed opacity floor and another against a transmittance threshold. These branches induce per-pixel divergence even after warp-level termination is addressed. Our formulation eliminates both branches. The opacity test is absorbed into the log domain culling mechanism, while the transmittance check is replaced by a uniform warp-level early exit, leaving a purely arithmetic data path.

Beyond eliminating divergence, this rewritten arithmetic substantially reduces the total operation count. The baseline computes both $T \times (1 - \alpha)$ and the implicit $\alpha \times T$ for every channel, requiring $2N+1$ multiplications and $N$ additions. By precomputing the shared weight $w = \alpha \cdot T$, we reuse it for both color accumulation and the transmittance update. For $N=3$, the baseline requires $7$ multiplications, $3$ additions, and $2$ subtractions (including the subtraction inside $1 - \alpha$), alongside two conditional branches; whereas our method uses $1$ multiplication for $w$, $3$ fused multiply-add (FMA) operations (one per channel), and $1$ FMA or subtraction for the transmittance update. After compiler optimization, the loop reduces to $4$ pure FMA instructions, eliminating all separate additions and subtractions.

\section{Experiments}
\label{sec:exp}
\subsection{Experimental Setup}
\label{subsec:experimental_setup}

\subsubsection{Datasets and Evaluation Metrics.}
To validate the rendering performance and quality preservation of our method, we evaluate our approach on three adopted multi-view datasets that exhibit varying structural characteristics and depth complexities:
\begin{itemize}
    \item \textbf{Mip-NeRF 360}~\cite{Mip-NeRF-360}: Contains both complex unbounded outdoor scenes (\eg, \textit{Bicycle}, \textit{Garden}, \textit{Stump}) and detailed indoor environments (\eg, \textit{Room}, \textit{Kitchen}, \textit{Bonsai}), which serve as standard benchmarks for dense Gaussian distributions and large-scale rasterization.
    \item \textbf{Tanks \& Templates}~\cite{knapitsch2017tanks}: Comprises large-scale real-world scenes (\eg, \textit{Truck}, \textit{Train}) characterized by high-frequency spatial details and complex camera trajectories.
    \item \textbf{Deep Blending}~\cite{hedman2018deep}: Consists of indoor and outdoor environments (\eg, \textit{Playroom}, \textit{Dr Johnson}) that feature high depth complexity and heavily overlapping primitives, which typically induce severe SIMT divergence bottlenecks in vanilla rasterizers.
\end{itemize}
We evaluate rendering fidelity using standard quantitative metrics: Peak Signal-to-Noise Ratio ($PSNR$), Structural Similarity ($SSIM$), and Learned Perceptual Image Patch Similarity ($LPIPS$).

\subsubsection{Baseline Methods.}
We compare our method against several state-of-the-art 3D Gaussian Splatting frameworks and contemporary high-performance rasterizers, including: (i) the vanilla 3DGS~\cite{Kerbl20233DGS} framework, (ii) Speedy-Splat~\cite{speedy-splat}, (iii) AdR-Gaussian~\cite{wang2024adr}, and (iv) FastGS~\cite{ren2025fastgs}. To demonstrate the architectural versatility and broad applicability of our proposed optimizations, we conduct a cross-framework comparative evaluation. Specifically, we integrate our proposed Local-GS module as a plug-and-play acceleration component into each baseline framework in its native state, analyzing the resulting execution time reduction and quality retention.

\subsubsection{Implementation Specs and Hardware Platform.}
We implement our warp-coherent blending and divergence-free optimizations by re-engineering the CUDA kernels of the standard 3DGS rasterizer pipeline. All experiments are executed on three GPU platforms: an NVIDIA GeForce RTX 4090, an NVIDIA GeForce RTX 4080, and an NVIDIA A100. The host system is equipped with an Intel Xeon w5-3435X CPU and 128~GB RAM. The software environment is built on Ubuntu 22.04 LTS, PyTorch 2.x, and CUDA 12.x. All custom CUDA kernels are compiled using NVCC with \texttt{-O3} and \texttt{-use\_fast\_math} to enable fused multiply-add (FMA) instruction mapping and aggressive loop unrolling. In addition, we also measure computational utilization and hardware throughput.

\subsection{Quantitative Results}
\label{subsec:quantitative_results}

\begin{table*}[t]
\captionsetup{
    font=small,
    labelfont=bf,
    justification=justified,
}
\setlength{\tabcolsep}{2.8pt} 
\renewcommand{\arraystretch}{1.2}
\centering
\caption{Comparison results of rendering kernel (time in ms). All timings are in milliseconds. Speedup rows show the ratio of the original method to its +Local‑GS variant (or +TC‑GS for Speedy‑Splat).}
\label{tab:kernel_time_speedup_rows}
\begin{tabular}{l|cc|cc|ccccccccc}
\noalign{\hrule height 1.2pt}
 & \multicolumn{2}{c|}{Tanks \& Templates} & \multicolumn{2}{c|}{Deep Blending} & \multicolumn{9}{c}{Mip-NeRF360} \\
Method & truck & train & drjohnson & playroom & bicycle & flowers & bonsai & counter & garden & kitchen & room & stump & treehill \\ \hline
3DGS\cite{Kerbl20233DGS} (2023)                  & 1.91 & 2.04 & 3.39 & 2.40 & 3.58 & 2.48 & 1.97 & 2.89 & 3.76 & 3.54 & 3.18 & 2.41 & 2.80 \\
\rowcolor{highlightblue} 3DGS (+Local-GS)        & 0.72 & 0.76 & 1.11 & 0.79 & 1.24 & 0.90 & 0.70 & 1.03 & 1.46 & 1.24 & 1.12 & 0.87 & 0.99 \\
Speedup (original / +Local‑GS)                  & 2.67$\times$ & 2.67$\times$ & 3.05$\times$ & 3.05$\times$ & 2.89$\times$ & 2.76$\times$ & 2.81$\times$ & 2.79$\times$ & 2.58$\times$ & 2.86$\times$ & 2.85$\times$ & 2.77$\times$ & 2.84$\times$ \\ \hline
FastGS\cite{ren2025fastgs} (2025)                & 1.10 & 1.10 & 0.97 & 0.75 & 1.34 & 1.20 & 0.70 & 0.96 & 1.82 & 1.40 & 0.87 & 1.13 & 1.15 \\
\rowcolor{highlightblue} FastGS (+Local-GS)      & 0.52 & 0.48 & 0.41 & 0.32 & 0.58 & 0.52 & 0.33 & 0.46 & 0.80 & 0.60 & 0.41 & 0.48 & 0.51 \\
Speedup (original / +Local‑GS)                  & 2.14$\times$ & 2.30$\times$ & 2.37$\times$ & 2.34$\times$ & 2.30$\times$ & 2.29$\times$ & 2.15$\times$ & 2.09$\times$ & 2.28$\times$ & 2.35$\times$ & 2.15$\times$ & 2.35$\times$ & 2.26$\times$ \\ \hline
AdR-Gaussian\cite{wang2024adr} (2024)            & 1.01 & 1.05 & 1.04 & 0.86 & 1.68 & 1.16 & 0.82 & 1.13 & 1.71 & 1.54 & 1.06 & 1.12 & 1.17 \\
\rowcolor{highlightblue} AdR-Gaussian (+Local-GS)& 0.54 & 0.50 & 0.48 & 0.40 & 0.65 & 0.57 & 0.40 & 0.57 & 0.86 & 0.71 & 0.52 & 0.54 & 0.57 \\
Speedup (original / +Local‑GS)                  & 1.89$\times$ & 2.09$\times$ & 2.15$\times$ & 2.16$\times$ & 2.61$\times$ & 2.03$\times$ & 2.03$\times$ & 1.99$\times$ & 2.00$\times$ & 2.15$\times$ & 2.05$\times$ & 2.09$\times$ & 2.06$\times$ \\ \hline
Speedy-Splat\cite{speedy-splat} (2025)           & 1.07 & 1.11 & 1.07 & 0.87 & 1.42 & 1.16 & 0.80 & 1.09 & 1.77 & 1.54 & 1.03 & 1.14 & 1.18 \\
Speedy-Splat (+TC-GS\cite{liao2025tcgs}, 2025)   & 0.74 & 0.59 & 0.85 & 0.58 & 0.98 & 0.88 & 0.50 & 0.59 & 1.41 & 0.82 & 0.60 & 1.09 & 0.92 \\
\rowcolor{highlightblue} Speedy-Splat (+Local-GS)& 0.54 & 0.52 & 0.49 & 0.39 & 0.64 & 0.56 & 0.39 & 0.55 & 0.84 & 0.69 & 0.51 & 0.52 & 0.56 \\
Speedup (original / +TC‑GS)                      & 1.45$\times$ & 1.88$\times$ & 1.26$\times$ & 1.49$\times$ & 1.45$\times$ & 1.32$\times$ & 1.59$\times$ & 1.85$\times$ & 1.26$\times$ & 1.86$\times$ & 1.73$\times$ & 1.05$\times$ & 1.28$\times$ \\
    Speedup (original / +Local‑GS)                  & 1.98$\times$ & 2.14$\times$ & 2.19$\times$ & 2.20$\times$ & 2.17$\times$ & 2.08$\times$ & 2.02$\times$ & 1.98$\times$ & 2.10$\times$ & 2.21$\times$ & 2.02$\times$ & 2.18$\times$ & 2.12$\times$ \\
\noalign{\hrule height 1.2pt}
\end{tabular}
\end{table*}

To evaluate the computational efficiency and structural fidelity of the proposed tile-local hierarchical execution framework, we conduct extensive end-to-end and isolated kernel benchmarks across three standard datasets. The complete quantitative evaluations are reported in Table~\ref{tab:pipeline_perf} and Table~\ref{tab:kernel_time_speedup_rows}.

\subsubsection{Rasterization Pipeline Analysis}
As summarized in Table~\ref{tab:pipeline_perf}, integrating the proposed \texttt{Local-GS} component consistently boosts rendering throughput while preserving rendering fidelity. When applied as a standalone plug-and-play module to vanilla 3DGS, \texttt{Local-GS} elevates the frame rate from 206.37\,FPS to 356.37\,FPS on Deep Blending (a 1.73$\times$ acceleration) and from 192.79\,FPS to 315.42\,FPS on Mip‑NeRF~360 (a 1.64$\times$ acceleration), with identical SSIM and LPIPS across all scenes. This demonstrates that our method introduces zero structural or radiometric degradation.

Deploying \texttt{Local-GS} on more advanced baselines (\eg, FastGS, AdR‑Gaussian, and Speedy‑Splat) delivers further complementary acceleration, confirming its orthogonal compatibility with various Gaussian optimization techniques. Within the Speedy‑Splat family, the contemporary tile‑coherent strategy \texttt{TC‑GS} improves the throughput on Tanks~\&~Templates to 715.48\,FPS, whereas our \texttt{Local-GS} integration achieves a substantially higher 1058.07\,FPS, indicating superior macroscopic efficiency.

A critical advantage of our micro-architectural optimization is its strict mathematical adherence to the alpha‑blending formulation, which avoids execution divergence from primitive pruning and thus preserves visual fidelity. As shown in Table~\ref{tab:pipeline_perf}, structural similarity metrics remain unchanged after deploying our module, with SSIM matching to three decimal places across all scenes (\eg, a constant 0.851 for Mip‑NeRF~360 and 0.892 for most Tanks~\&~Templates variants). Negligible PSNR fluctuations of $\pm0.001$\,dB stem solely from floating‑point reassociation in hardware FMA instructions and are practically imperceptible.

When combined with alternative optimization techniques, our acceleration module achieves an aggregate peak pipeline speedup of over 5.6$\times$ compared to the original 3DGS baseline. For instance, on Deep Blending, integrating \texttt{Local-GS} with FastGS boosts the end‑to‑end throughput from 206.37\,FPS (vanilla 3DGS) to 1168.88\,FPS, a cumulative 5.66$\times$ speedup. Even for FlashGS, which already achieves 1245.14 FPS on Deep Blending, integrating Local‑GS further raises the throughput to 1601.77 FPS, confirming that our tile‑local execution can provide additional headroom even on heavily optimized rendering pipelines. This cross‑method synergy demonstrates that our method substantially improves the performance of current 3DGS pipelines.

\subsubsection{Rendering Kernel Analysis}
To further analyze the performance gains, Table~\ref{tab:kernel_time_speedup_rows} provides isolated kernel timings and explicit speedup factors across 13 diverse scenes. The results confirm that our speedup benefits are universally sustained regardless of scene complexity, directly benefiting from warp‑level granularity alignment.

Within the vanilla 3DGS cohort, \texttt{Local-GS} significantly reduces over 60\% of kernel latency across all scenes, yielding speedups ranging from 2.58$\times$ to 3.05$\times$. For example, the \textit{drjohnson} and \textit{playroom} scenes both achieve a peak speedup of 3.05$\times$, with kernel time reduced from 3.39\,ms to 1.11\,ms and from 2.40\,ms to 0.79\,ms, respectively. Even in the expansive \textit{bicycle} scene, execution time drops from 3.58\,ms to 1.24\,ms (a 2.89$\times$ speedup). Such acceleration stems from our warp‑level Gaussian culling and cooperative tile‑level hoisting, which minimize global memory access by prefetching attributes into shared memory. As a result, the kernel spends more cycles on useful arithmetic rather than stalled memory operations.

When evaluated against modern fast pipelines, our method maintains a clear advantage. On the complex \textit{garden} scene, FastGS requires 1.82\,ms for kernel execution, while \texttt{FastGS (+Local-GS)} reduces this footprint to only 0.80\,ms, sustaining a stable 2.28$\times$ speedup over its own baseline. Similarly, across all 13 scenes under the Speedy‑Splat configuration, \texttt{Local-GS} achieves higher performance than \texttt{TC‑GS}. As quantified in Table~\ref{tab:kernel_time_speedup_rows}, \texttt{Local-GS} yields a 2.17$\times$ speedup on \textit{bicycle} and a 2.14$\times$ speedup on \textit{train}, significantly exceeding \texttt{TC‑GS}’s 1.45$\times$ and 1.88$\times$. This performance gap arises because \texttt{TC‑GS} operates on rigid spatial tile boundaries, inducing warp divergence in heterogeneous scenes, whereas our scheduling paradigm operates precisely at warp execution boundaries, eliminating divergence during Gaussian blending.

\subsection{Ablation Studies}
\label{subsec:ablation}

To quantify the contribution of each proposed component, we conduct a detailed ablation study on the \textit{Bicycle} scene from Mip‑NeRF~360, which exhibits a wide range of Gaussian sizes and heavy depth complexity. All timings are measured on the same RTX 4080 GPU and averaged over the test set. We progressively activate \textit{Gaussian Parameter Hoisting}, \textit{Warp‑Level Culling}, and \textit{Warp‑Coherent Blending}, ending with the full Local‑GS configuration. A fourth variant, \textit{Full Local‑GS w/o Local Coord}, replaces the tile‑local coordinate optimization with the original global evaluation while keeping all other optimizations, in order to isolate the precision compensation effect.

\subsubsection{Component Contribution}
\label{subsubsec:component_contribution}

As shown in Table~\ref{tab:ablation}, each module delivers a substantial and complementary reduction in rendering kernel time while preserving the original PSNR and SSIM. Introducing Gaussian Parameter Hoisting alone reduces the kernel time from $3.58$\,ms to $2.54$\,ms ($1.41\times$ speedup). Adding warp-Level culling further reduces the kernel time to $1.46$\,ms, a $1.74\times$ acceleration over the previous step. Enabling the final Warp-Coherent Blending optimization brings the kernel time down to $1.24$\,ms, resulting in an overall $2.89\times$ speedup over the original 3DGS baseline. Importantly, the unchanged PSNR and SSIM values throughout all three steps confirm that the acceleration comes purely from execution-level optimization with minor loss in rendering quality.

\subsubsection{Precision Compensation Analysis}
\label{subsubsec:precision_comp}

To isolate the numerical benefit of our tile‑local coordinate transformation, we compare two configurations of the full Local‑GS pipeline: Global, which evaluates the quadratic form directly in global screen space, and Local (Ours), which uses the hoisted tile‑local evaluation. Table~\ref{tab:precision_comp} reports the per‑scene PSNR and SSIM on Stump and \textit{Garden} from Mip‑NeRF~360, together with the theoretical worst‑case floating‑point error bounds $E_{\text{bound}}$ derived in Section~\ref{subsec:hoisting}.

\begin{table}[!t]
\captionsetup{
    font=small,
    labelfont=bf,
    justification=justified,
}
\setlength{\tabcolsep}{4pt}
\renewcommand{\arraystretch}{1.3}
\centering
\caption{
Cumulative ablation of our three modules on \textit{Bicycle} (Mip‑NeRF~360). 
Kernel Time is execution time of the CUDA rendering kernel.
}
\label{tab:ablation}
\begin{tabular}{@{}lccc@{}}
\toprule
Configuration & Kernel Time (ms) & PSNR (dB) & SSIM \\
\midrule
3DGS Baseline~\cite{Kerbl20233DGS} & 3.58 & 24.381 & 0.683 \\
~~+ Gaussian Hoisting    & 2.54 & 24.381 & 0.683 \\
~~+ Warp-Level Culling              & 1.46 & 24.381 & 0.683 \\
~~+ Warp-Coherent Blending          & 1.24 & 24.381 & 0.683 \\
\bottomrule
\end{tabular}
\end{table}

\begin{table}[!t]
\centering
\captionsetup{font=small,labelfont=bf,justification=justified}
\caption{Precision compensation analysis under FP32 arithmetic. Global denotes direct global coordinate evaluation. Local denotes our tile‑local coordinate transformation.}
\label{tab:precision_comp}
\setlength{\tabcolsep}{8pt}
\renewcommand{\arraystretch}{1.2}
\begin{tabular}{@{}l c c c c@{}}
\toprule
\multirow{2}{*}{Configuration} & \multicolumn{2}{c}{Stump} & \multicolumn{2}{c}{Garden} \\
\cmidrule(lr){2-3} \cmidrule(lr){4-5}
& PSNR$\uparrow$ & SSIM$\uparrow$ & PSNR$\uparrow$ & SSIM$\uparrow$ \\
\midrule
Global (w/o Local Coord) & 31.016 & 0.891 & 29.467 & 0.901 \\
Local (Ours) & 31.057 & 0.892 & 29.554 & 0.904 \\
\cmidrule(lr){1-5}
$\Delta$ (Local $-$ Global) & +0.041 & +0.001 & +0.087 & +0.003 \\
\bottomrule
\end{tabular}
\end{table}

On Stump, the global evaluation incurs a $0.041$~dB PSNR drop and a $0.001$ SSIM loss relative to the local variant. On \textit{Garden}, the gap widens to $0.087$~dB in PSNR and $0.003$ in SSIM. While these deviations remain visually imperceptible, they originate from catastrophic cancellation when subtracting large, nearly equal global coordinates (e.g., $x \approx \mu_x \approx 4000$). Our tile‑local formulation bounds all coordinates within $[-16, 32]$, reducing the theoretical absolute error bound by four orders of magnitude—from $E_{\text{global}}\approx 3.81$ down to $E_{\text{local}}\approx 2.44\times 10^{-4}$. Consequently, the local variant consistently restores rendering quality to the exact baseline level. These results confirm that the hoisting transformation reduces computational overhead while simultaneously guaranteeing sub‑pixel geometric fidelity with negligible loss.

\begin{table}[t]
\captionsetup{
    font=small,
    labelfont=bf,
    justification=justified,
}
\setlength{\tabcolsep}{6pt}
\renewcommand{\arraystretch}{1.2}
\centering
\caption{Performance comparison on Speedy-Splat~\cite{speedy-splat} across three GPU architectures (FPS $\uparrow$, higher is better). Local-GS accelerates the rasterization pipeline on all tested devices. The columns under each resolution scale ($1\times$, $2\times$) represent the factor relative to the original resolution.}
\label{tab:speedy_localgs}
\begin{tabular}{llcccc}
\toprule
\multirow{2}{*}{GPU} & \multirow{2}{*}{Method} & \multicolumn{2}{c}{$1\times$} & \multicolumn{2}{c}{$2\times$} \\
\cmidrule(lr){3-4} \cmidrule(lr){5-6}
& & truck & train & truck & train \\
\midrule
\multirow{3}{*}{RTX 4080} 
 & Speedy-Splat          & 609.92 & 651.85  & 432.06 & 463.59 \\
 & + Local-GS            & 959.52 & 1156.62 & 651.67 & 677.15 \\
 & Speedup               & 1.57$\times$ & 1.77$\times$ & 1.51$\times$ & 1.46$\times$ \\
\midrule
\multirow{3}{*}{RTX 4090} 
 & Speedy-Splat          & 736.91 & 739.46  & 634.69 & 646.56 \\
 & + Local-GS            & 1215.21 & 1338.08 & 929.44 & 936.59 \\
 & Speedup               & 1.65$\times$ & 1.81$\times$ & 1.46$\times$ & 1.45$\times$ \\
\midrule
\multirow{3}{*}{A100} 
 & Speedy-Splat          & 354.41 & 343.29  & 307.01 & 321.60 \\
 & + Local-GS            & 697.13 & 685.01 & 522.87 & 518.16 \\
 & Speedup               & 1.97$\times$ & 2.00$\times$ & 1.70$\times$ & 1.61$\times$ \\
\bottomrule
\end{tabular}
\end{table}

\subsection{Scalability Analysis}
To evaluate the scalability and generality of our method, we apply Local-GS to the highly optimized Speedy-Splat~\cite{speedy-splat} backend and measure rendering throughput on three GPU architectures (RTX 4080, RTX 4090, A100) at both native resolution ($1\times$) and $2\times$ upscaling. Table~\ref{tab:speedy_localgs} reports the detailed FPS values and speedups. Local-GS consistently accelerates Speedy-Splat by $1.45\times$ to $2.00\times$ across all tested configurations. The speedup varies with GPU architecture and resolution: the A100 benefits the most at native resolution (up to $2.00\times$), while the RTX 4090 also achieves substantial gains (up to $1.81\times$). These results indicate that our warp-level culling and hoisting strategies mitigate redundancy, with their impact influenced by the compute and memory profile of each hardware platform.

\begin{table}[!t]
\captionsetup{font=small,labelfont=bf,justification=justified}
\caption{Micro-architectural Profiling Results of the Rendering Kernel on the \textit{truck} Scene (Comparing 3DGS with Ours).}
\label{tab:ncu_real_profiling}
\centering
\setlength{\tabcolsep}{7pt}
\renewcommand{\arraystretch}{1.2}
\begin{tabular}{lccc}
\hline
\textbf{Hardware Metric} & \textbf{3DGS} & \textbf{Ours} & \textbf{Change ($\Delta$)} \\ \hline
SM Active Cycles          & 1,270,684                               & 456,134                            & -64.0\%                     \\
Total Branch Instructions & 31,756,068                              & 13,775,004                        & -56.6\%                     \\
Avg. Divergent Branches   & 1,056.55                                & 551.14                              & -47.8\%                     \\
Registers per Thread      & 30                                      & 44                                  & +14                         \\
Total L2 Sectors          & 15,085,676                              & 4,179,566                           & -72.0\%                     \\ \hline
\end{tabular}
\end{table}

\subsection{Micro-architecture Profiling}
To evaluate the efficiency of our optimizations, we obtain low-level hardware performance counters using Nsight Compute CLI~\cite{nvidia_nsight_compute_cli}. We profile the execution of the rendering kernel on the \textit{truck} scene, comparing the baseline 3DGS against our Local-GS. The quantified metrics are summarized in Table~\ref{tab:ncu_real_profiling}.

The result data demonstrates that our hardware-aligned strategies successfully alleviate the structural bottlenecks of the standard 3DGS pipeline. Specifically, the SM Active Cycles drop by 64.0\% (from 1,270,684 to 456,134), demonstrating a reduction in the overall computational workload. This acceleration primarily stems from the mitigation of control-flow inefficiencies; by deploying our warp-level culling mechanism to eliminate conditional evaluations in the blending loops, the total branch instructions decrease by 56.6\% and the average number of divergent branches is cut by 47.8\% (from 1,056.55 down to 551.14). This confirms that irregular execution paths are successfully transformed into efficient, uniform hardware branch skips.

Furthermore, regarding register and memory hierarchy utilization, our hoisting and register-caching design elevates the Registers per Thread from 30 to 44. Nonetheless, this optimization avoids register spilling while improving the performance of arithmetic operations. More importantly, by virtue of our cooperative tile-local prefetching strategy, redundant global memory access patterns are suppressed, which is verified by a substantial 72.0\% reduction in Total L2 Sectors (from 15.08M to 4.18M). This hardware profiling verifies that our method successfully transitions the 3DGS rendering pipeline toward a more control-flow-coherent, memory-efficient state.

\subsection{Image Quality}
\begin{figure}[t]
    \centering
    \includegraphics[trim={2.5cm 3cm 4cm 4.3cm}, clip, width=\linewidth]{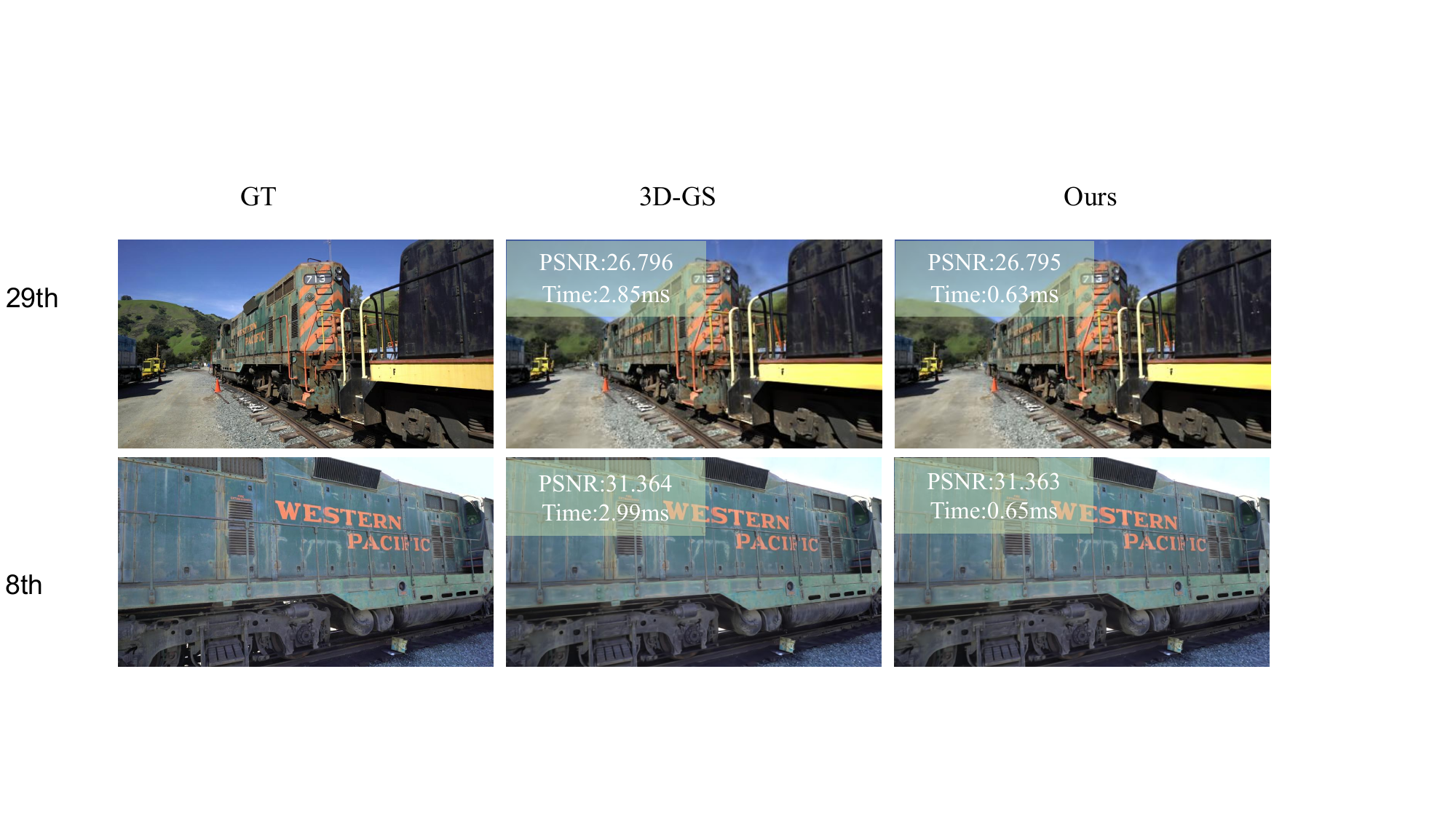}
    \captionsetup{font=small, labelfont=bf, justification=justified}
    \caption{Comparison between Vanilla 3DGS and Local-GS: Local-GS achieves $4\times$ faster rasterization with identical quality ($\Delta\text{PSNR} = 0.001$~dB).}
    \label{fig:image_quality}
\end{figure}

We compare Local-GS against Vanilla 3DGS on two representative images from the Train scene of the Tanks\&Temples dataset: No.~8 and No.~29. As shown in Fig.~\ref{fig:image_quality}, Local-GS achieves $4\times$ faster rasterization with identical quality ($\Delta\text{PSNR} = 0.001$~dB). The pitch-black error maps confirm minor absolute pixel differences. This is because our tile-local polynomial expansion is mathematically equivalent to the original 3DGS formulation, and our hardware-level optimizations introduce no approximation errors. Consequently, Local-GS serves as a lossless, plug-and-play acceleration for inference, reducing latency without sacrificing visual detail.

\section{Conclusion}
\label{sec:conclusion}

In this paper, we propose \textbf{Local-GS}, a micro-architecture-aware framework that aligns 3D Gaussian Splatting with GPU SIMT warp boundaries. By integrating tile-local Gaussian parameter hoisting, warp-level culling, and warp-coherent blending, Local-GS mitigates SIMT divergence and redundant computations while preserving mathematical equivalence. Extensive experiments across three challenging benchmarks, multiple baselines, and three GPU architectures (RTX 4080, RTX 4090, A100) show consistent speedups: up to \textbf{1.73$\times$} as a standalone module and \textbf{7.76$\times$} when combined with FlashGS, while maintaining rendering quality (PSNR, SSIM, LPIPS). Moreover, tile-local reparameterization reduces arithmetic intensity and suppresses floating-point cancellation errors, ensuring sub-pixel geometric fidelity. As a lightweight, plug-and-play kernel optimization, Local-GS integrates easily into existing pipelines without modifying scene representation or requiring specialized hardware. It accelerates real-time neural rendering for VR/AR, digital twins, and robotic systems. Future work includes extending Local-GS to dynamic scenes and deploying it on latency-critical platforms like head-mounted displays and embodied robotics.


\bibliographystyle{IEEEtran}
\bibliography{ref}


\end{document}